\pdfoutput=1

\documentclass[letterpaper, 10 pt, conference]{ieeeconf}  

\IEEEoverridecommandlockouts                              

\title{\LARGE \bf
Instance-Level Semantic Maps for Vision Language Navigation
}

\author{Laksh Nanwani$^{1}$, Anmol Agarwal$^{1}$, Kanishk Jain$^{1}$, Raghav Prabhakar$^{1}$, \\Aaron Monis$^{1}$, Aditya Mathur$^{1}$, Krishna Murthy Jatavallabhula$^{2}$,\\A. H. Abdul Hafez$^{3}$, Vineet Gandhi$^{1}$, K. Madhava Krishna$^{1}$
\thanks{$^{1}$KCIS, International Institute of Information Technology, Hyderabad, India.}
\thanks{$^{2}$CSAIL, MIT, Cambridge, United States.}
\thanks{$^{3}$Hasan Kalyoncu University, Sahinbey, Gaziantep, Turkey.}
}

\makeatletter
\def\@maketitle{\newpage
\null
\begin{center}%
\let \footnote \thanks
{\LARGE \@title \par}%
\vskip 1.5em%
{\large
\lineskip .5em%
\begin{tabular}[t]{c}%
\@author
\end{tabular}\par}%
\end{center}%
\vskip -5em
\par
}
\makeatother

\usepackage{graphics} 
\usepackage{epsfig} 
\usepackage{mathptmx} 
\usepackage{times} 
\usepackage{amsmath} 
\usepackage{amssymb}  
\usepackage{booktabs}
\usepackage{multirow}
\usepackage{cuted}
\usepackage{stfloats}
\usepackage{caption}
\usepackage{todonotes}
\usepackage{xcolor}
\usepackage[margin=0.75in]{geometry}
\usepackage{url}

\begin{document}

\maketitle
\thispagestyle{empty}
\pagestyle{empty}

\vspace{-10cm}
\setlength{\belowcaptionskip}{-8pt}

\begin{strip}
  \centering
  \includegraphics[width=0.9\linewidth]{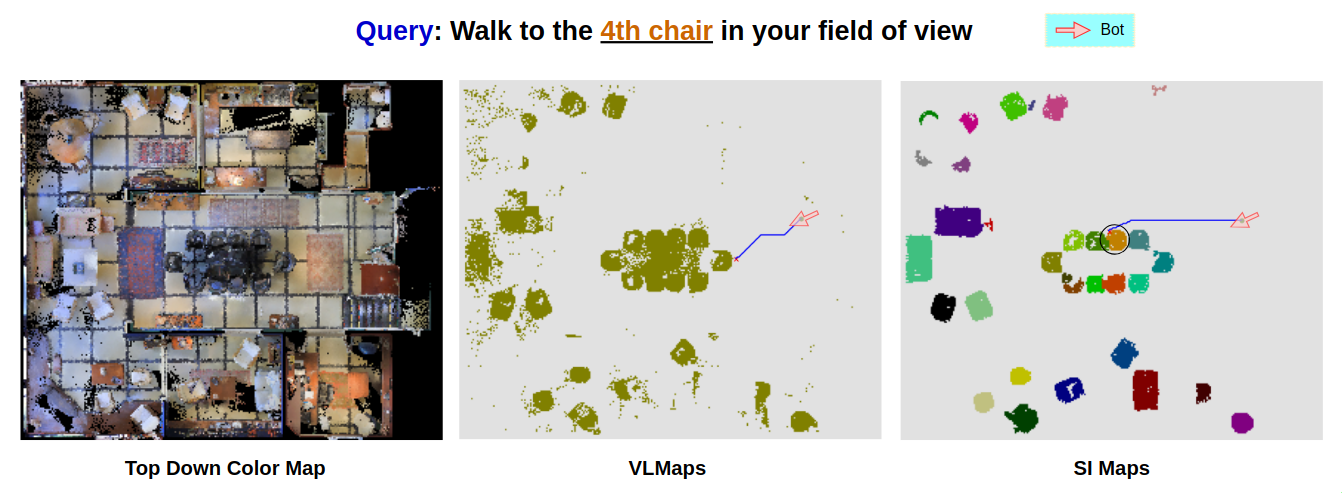}
  \captionof{figure}{VLMaps \cite{huang23vlmaps}, being a semantic top-view map, cannot distinguish between different instances of the same object. On the other hand, SI Maps (ours) are directly amenable for handling such queries as they contain instance-specific information for all objects in the environment. For the scene on the extreme left, the instances of the object `chair' as detected by SI Maps is shown in different colors in the rightmost figure.}
  \label{fig:figure1}
\end{strip}
\begin{abstract}

Humans have a natural ability to perform semantic associations with the surrounding objects in the environment. This allows them to create a mental map of the environment, allowing them to navigate on-demand when given linguistic instructions. A natural goal in Vision Language Navigation (VLN) research is to impart autonomous agents with similar capabilities. Recent works take a step towards this goal by creating a semantic spatial map representation of the environment without any labeled data. However, their representations are limited for practical applicability as they do not distinguish between different instances of the same object. In this work, we address this limitation by integrating instance-level information into spatial map representation using a community detection algorithm and utilizing word ontology learned by large language models (LLMs) to perform open-set semantic associations in the mapping representation. The resulting map representation improves the navigation performance by two-fold (233\%) on realistic language commands with instance-specific descriptions compared to the baseline. We validate the practicality and effectiveness of our approach through extensive qualitative and quantitative experiments.

\end{abstract}

\section{INTRODUCTION}

Advancements in machine learning research have brought about rapid changes in the field of robotics, allowing for the development of sophisticated autonomous agents. However, making this technology practically viable for large-scale adoption requires a natural mechanism to interact with humans. Vision Language Navigation (VLN) research aims to achieve this goal by incorporating natural language understanding into autonomous agents to navigate the environment based on linguistic commands. Prior approaches to VLN have addressed this task by harnessing the capabilities of visual grounding models, which allow the navigating agents to localize objects in the visual scene or directly ground navigable regions based on linguistic descriptions. However, these approaches fail to address linguistic commands which require spatial precision to identify the goal region. Furthermore, these approaches assume that the object referred to by the linguistic command is always visible in the current scene. Such an assumption rarely holds in realistic scenarios, where things can move in or out of the current scene as we navigate the environment. 

Consider the example in Figure \ref{fig:figure1} with the language command, ``walk to the fourth chair in your field of view". To execute this command, we first need to explore the entire room to find all instances of chairs and then find the fourth instance from where the command was given. For visual grounding-based approaches, it is non-trivial to handle such scenarios as there is no way to rank the localized chairs based on distance. To counteract the above issues, geometric maps, which create a global mapping of the surrounding environment, provide a direct mechanism to ground all the objects present in the scene, including those not visible in the current view, and additionally, are readily amenable for planning and navigation purposes. In this work, we propose a memory-efficient mechanism for creating a semantic spatial representation of the environment, which is directly applicable to robots navigating in real-world scenes.

\begin{figure*}[ht]
  \centering
  \includegraphics[width=0.88\linewidth]{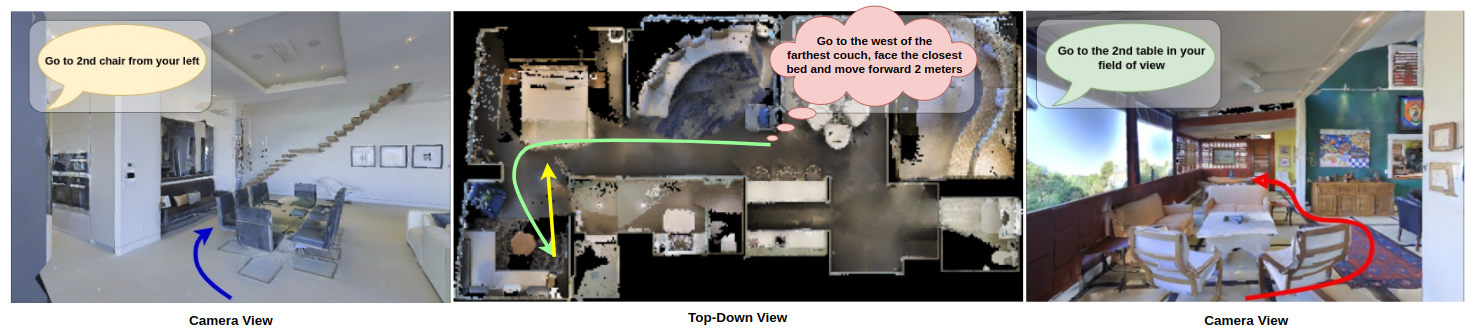}
  \caption{Our top-view map representation allows indoor embodied agents to perform complex instance-specific goal navigation in object-rich environments. The language queries can refer to individual instances based on spatial and viewpoint configuration with respect to other objects of the same type while preserving the navigation performance on standard language queries.}
  \label{fig:figure2}
  \vspace{-1em}
\end{figure*}

Recent works like VLMaps and NLMap \cite{chen2022nlmapsaycan} propose a mechanism to build semantic spatial maps without any labeled data by fusing pre-trained vision-language features with the 3D point cloud of the physical world. They compute the similarity between visual and linguistic features in a common semantic space of a large-scale pre-trained vision-language model and utilize large-language models to convert the natural language command to a sequence of navigation goals for planning. However, their map representation doesn't allow them to differentiate between different instances of the same object and hence handle language queries that describe an instance-specific navigation goal, like the ones mentioned in Figure \ref{fig:figure2}, as the visual encodings are instance-agnostic. Moreover, their mechanism is memory intensive as they require high-dimensional feature embeddings to make semantic associations for the objects in the visual scene.

Our work focuses on creating spatial maps of the environment with instance-level semantics. We achieve this in a memory-efficient manner, bypassing the use of feature embeddings altogether. We show that \textbf{S}emantic \textbf{I}nstance Maps (SI Maps) are computationally efficient to construct and allow for a wide range of complex and realistic commands that evade prior works.

\section{RELATED WORK}

\subsection{Semantic Mapping}
With the recent progress in computer vision and natural language processing literature, there has been considerable interest in augmenting the semantic understanding of traditional SLAM algorithms. Earlier works like SLAM++ \cite{salas2013slam++} propose an object-oriented SLAM, which utilizes prior knowledge about the domain-specific objects and structures in the environment. Later works like \cite{mccormac2018fusion++} assign instance-level semantics using Mask-RCNN \cite{he2017mask} to 3D volumetric maps. Some methods \cite{huang23vlmaps, chen2022nlmapsaycan} have also explored transferring predictions from CNNs in 2D pixel space to 3D space for 3D reconstruction. Concurrent to our work, \cite{AyushMultiON} proposes a deep reinforcement learning-based approach for multi-object instance navigation, albeit without linguistic commands. VLMaps \cite{huang23vlmaps} and NLMap-Saycan \cite{chen2022nlmapsaycan} propose a natural language queryable scene representation with Visual Language models (VLMs). These methods utilize large-language models (LLMs) to parse language instructions and identify the involved objects to query the scene representation for object availability and location.

\subsection{Instance Segmentation}
The ability to identify and localize different instances of similar objects is crucial for visual perception tasks in robotics. In the Computer Vision literature, the task of instance segmentation serves to evaluate such capabilities formally. Earlier works \cite{he2017mask} utilized region proposal networks to predict candidate bounding boxes followed by a mask head to regress the instance-level segmentation mask for each proposal. While initial approaches designed task-specific architectures, more recent methods \cite{cheng2021mask2former} have moved towards generalized architectures for different image segmentation tasks like semantic, instance, and panoptic segmentation. Mask2Former \cite{cheng2021mask2former} employs attention mechanism to extract localized object-centric features in an end-end manner. In this work, we utilize segmentation masks from Mask2Former to create instance-level semantic maps which are directly amenable for planning during autonomous navigation.

\subsection{Vision Language Navigation}

Most of the work in Vision Language Navigation (VLN) has focused on navigating in the environment using semantic perception based on the front camera view of the autonomous agent. Specifically, these works take the front camera image and the language command as input, and the navigation task is reduced to a sequence modeling task where at each time stamp, the optimal action is predicted to complete the navigation task successfully. Subsequent works have tackled the VLN problem using sequence-to-sequence learning \cite{schumann-riezler-2022-analyzing}, reinforcement learning \cite{nguyen2019vision} or behavior cloning methods \cite{das2018neural}. However, these methods are non-trivial to interpret, and recent works \cite{schumann-riezler-2022-analyzing} have found that such methods are unable to utilize the visual modality effectively for the navigation task. Consequently, recent works \cite{huang23vlmaps, chen2022nlmapsaycan} on VLN have focused on creating a semantic map of the environment for motion planning and utilizing visual grounding capabilities of large-scale vision-language models \cite{radford2021learning} to ground the semantic concepts in a visual world. In this work, we focus on creating a semantic mapping representation of the environment using large-scale language models. Unlike prior works, we create these maps in an embedding-free manner, thus reducing the computational cost significantly.

\section{Method}

\subsection{Problem Statement}
\begin{figure*}[ht]
  \centering   
  \includegraphics[width=0.8\textwidth]{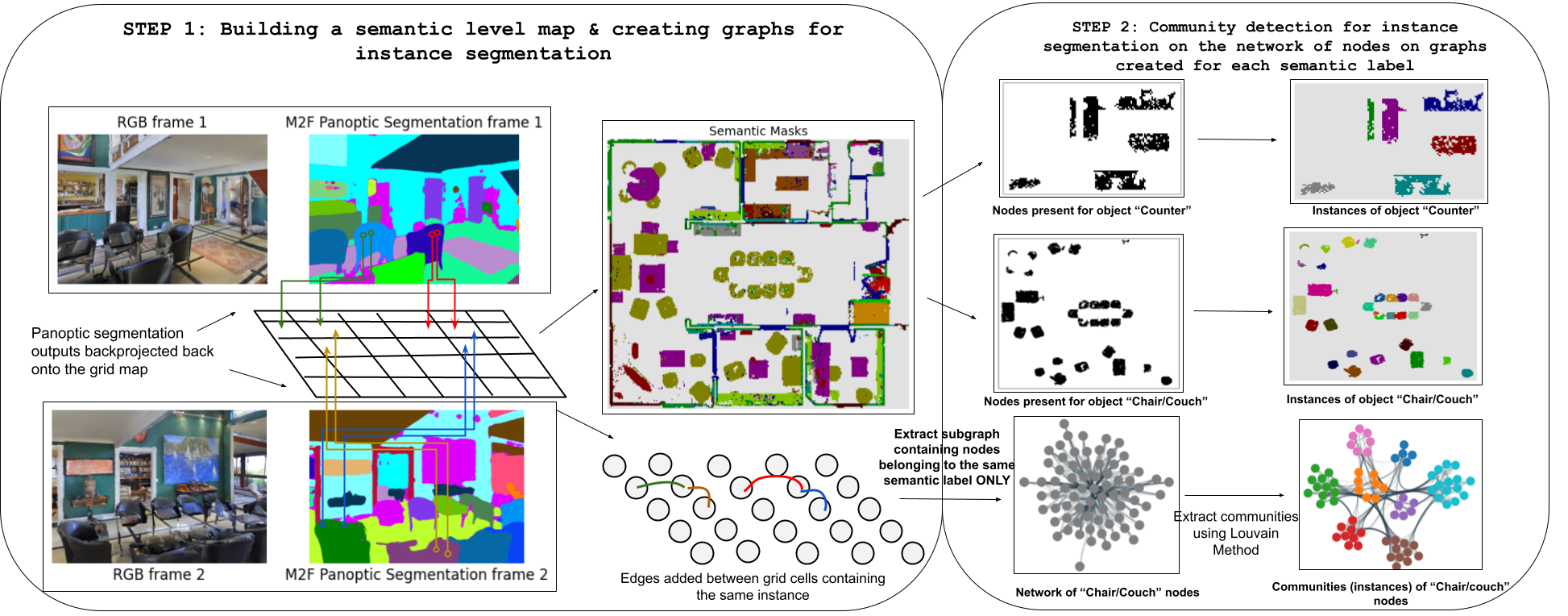} 
  \caption{In STEP 1, we create a semantic level map of the environment by back projecting the Mask2Former semantic labels of the RGB pixels across different images onto the grid map. In STEP 2, we extract the subgraph concerned with object $o$ and run a community detection algorithm to break the grid cells containing object $o$ into instances. 
  }
  \label{fig:figure3}
\end{figure*}

In this work, we aim to create a semantic map of the surrounding environment containing instance-level information for the various objects. Maps containing both instance-level and semantic information are necessary to handle linguistic commands which are frequently used in the daily vernacular. For example, consider the command, ``Go to the empty chair near the third table". We are required to identify ``which instance of the table" is being talked about and then point out the instance of the empty chair. Our approach is equipped to handle such scenarios through an instance-specific mapping representation of the environment. We build SI Maps using only RGB-D sensors, pose information, and an off-the-shelf panoptic segmentation model. SI Maps creation involves two steps: (1) Occupancy map creation with semantic labels and (2) Community detection to separate instances of a given semantic label. The whole pipeline is illustrated in Figure \ref{fig:figure3}.

\subsection{SI Map Creation}
\textbf{Building Occupancy Grid: } We define SI Maps as $\mathcal{M} \in \mathbb{R}^{\bar{H} \times \bar{W} \times 2}$, where $\bar{H}$ and $\bar{W}$ represent the size of the top-down grid map. Similar to VLMaps, with the scale parameter $s$ ($=0.05 m$ in our experiments), a SI Map $\mathcal{M}$ represents an area with size $s\bar{H} \times s\bar{W}$ square meters. $\mathcal{M}_{i,j} = \textlangle o, t \textrangle$ means that grid cell $(i, j)$ is occupied by the $t^{th}$ instance of object $o$ in the environment. Since we are using the Mask2Former panoptic segmentation model trained on the COCO dataset \cite{lin2014microsoft}, $o \in \mathcal{O}$ (where $\mathcal{O}$ is the set of objects present in the COCO dataset). To build our map, similar to VLMaps, we, for each RGB-D frame, back-project all the depth pixels $\mathbf{u} = (u, v)$ to form a local depth point cloud that we transform to the world frame using the pose information. For depth pixel $\mathbf{u} = (u, v)$ belonging to the $i^{th}$ RGB-D frame, let $(p^{i, u,v}_{x},\:p^{i, u,v}_{y})$ represent the coordinates of the projected point in the grid map $\mathcal{M}$. 

\textbf{Integrating Instance-level information: } With the occupancy map defined, we now utilize community detection algorithms to separate out the different instances in the environment. Specifically, we use the modularity-based Louvain method, a greedy, hierarchical optimization method that iteratively refines communities to maximize the modularity value. The modularity value is a measure of the density of links within communities compared to links between communities. 

Let the output of the panoptic segmentation model for $\mathbf{u} = (u, v)$ be $\langle o_{i, u, v}, t_{i, u, v} \rangle$. This means object $o_{i, u, v}$'s $t_{i, u, v}^{th}$ instance within the frame is present at pixel $\mathbf{u}$. We use this information to set the object label $o$ for $\mathcal{M}_{(p^{i, u,v}_{x},p^{i, u,v}_{y})} $ as $o_{i, u, v}$. When there exist multiple 3D depth pixels projecting to the same grid location in the map, we retain the label of the pixel with the highest vertical height. 

To divide the different grid cells labeled having object $o$ into different instances, we construct an undirected weighted graph $G = (V, E, W)$, where each grid cell $(i, j)$ for whom the object label of $\mathcal{M}_{i,j}$ is equal to $o$ is included as a node in the set of vertices $V$. Whenever two neighbouring pixels $\mathbf{u_1} = (u_1, v_1)$ and $\mathbf{u_2} = (u_2, v_2)$ belong to the same entity in the $i^{th}$ RGB-D frame, their corresponding grid cells $(p^{i, u_1,v_1}_{x},p^{i, u_1,v_1}_{y})$ and $(p^{i, u_2,v_2}_{x},p^{i, u_2,v_2}_{y})$ should also belong to the same instance in real-world. Hence, whenever pixels $\mathbf{u_1}$ and $\mathbf{u_2}$ have the semantic label $o$ and $\textlangle o_{i, u_1, v_1}, t_{i, u_1, v_1} \rangle = \langle o_{i, u_2, v_2}, t_{i, u_2, v_2} \rangle$, i.e., depth pixels $\mathbf{u_1}$ and $\mathbf{u_2}$ belong to the same entity within the image, we increase the edge weight between grid cells $(p^{i, u_1,v_1}_{x},p^{i, u_1,v_1}_{y})$ and $(p^{i, u_2,v_2}_{x},p^{i, u_2,v_2}_{y})$ by one. This helps us in transferring the instance segmentation information present in the panoptic segmentation outputs of the RGB-D frames to our map and also helps us to track the same instance across frames using the pose data. To prevent the frequency of visiting a particular area in the environment during mapping from unfairly affecting any edge weight, we normalize all the edge weights by the number of times their constituent nodes (grid cells) were observed across all RGB-D images for that scene. Ideally, in our graph, all grid cells belonging to the same connected component should belong to the same real-world entity. But Mask2Former masks are not perfect at a pixel level; hence it is possible for spurious edges to be drawn between nodes belonging to different real-world entities. However, such edges are likely to be few in number. To disregard such spurious edges, we group the nodes in $V$ using community detection algorithms instead of naively breaking them into connected components.\\
We initialize the graph with a separate community for each node. We use the Louvain community detection method, which involves two phases: (1) Modularity optimization and (2) Community aggregation. During modularity optimization, for each node in the graph, we compute the change in modularity by moving it to neighboring communities. The node is transferred to the community, which results in the highest increase in modularity. This procedure is repeated for all nodes until no further improvement in modularity is possible. In the community aggregation phase, the communities formed in the modularity optimization phase are considered single nodes. The weights of the edges between the new nodes are determined by the sum of the weights of the edges between the nodes in the original communities. The two phases are iteratively repeated until the modularity value converges. After convergence, we get a labeled graph, where the nodes are grouped based on their community membership, i.e., occupancy grid cells belonging to the same instance are grouped together for all the objects in the environment. To correct the over-segmentation of communities, a post-processing step is applied to merge communities ${C_1}$ and ${C_2}$ if more than $K\%$ of the members of ${C_1}$ are neighbors of some member of ${C_2}$.

In contrast to VLMaps, our approach doesn't utilize the high dimensional LSeg \cite{li2022languagedriven} feature embeddings for semantic map creation, which provides a memory-efficient mechanism to construct the instance-level semantic occupancy grid. For comparison, VLMaps representation requires an average storage of about 2 gigabytes for a $1000 \times 1000$ map, whereas SI Maps needs only about 16 megabytes for the same map size. Additionally, the proposed approach is highly flexible and adaptable, as it can easily incorporate other types of sensor data like LiDar, IMU and plug different segmentation models. The provision of tunable hyper-parameter $K$ further provides controllability in our approach, which is a desired capability for real-world deployment. In the next section, we show how SI Maps can be directly used for language-conditioned navigation.

\subsection{Language-based Navigation}

The significance of Semantic Instance maps becomes apparent when dealing with commands that necessitate instance-level grounding. For a given language command, we would like to identify the region in SI Maps where the robot must navigate to execute the command successfully. Additionally, since different commands can refer to different navigational maneuvers, we must also determine the maneuvers required for a specific language query. To achieve this, we define function primitives for each possible maneuver, reducing the task to classifying the appropriate function primitive for each sub-command. For this classification, we utilize the powerful large language model (LLM), ChatGPT\cite{OpenAI}, for motion planning.

\begin{figure}[ht]
  \centering
  \includegraphics[width=\columnwidth]{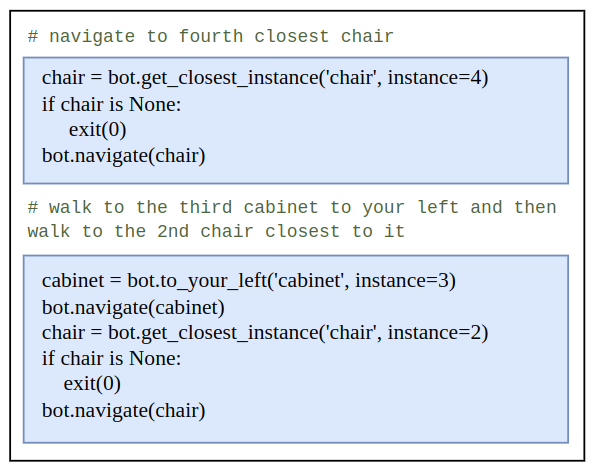}
  \caption{An example of the executable Python code generated by ChatGPT for the given language commands. The generated code includes an instance parameter in the function primitive call for navigating to the specified instance in the environment.}
  \label{fig:figure4}
  \vspace{-0.9em}
\end{figure}

LLMs, trained on billions of lines of text and code, demonstrate advanced natural language understanding, reasoning, and coding capabilities. Similar to the approach with VLMaps, we repurpose LLMs to generate executable Python code for the robot. Specifically, we supply ChatGPT with the list of function primitives and their respective descriptions. We then prompt ChatGPT with several language queries accompanied by the corresponding ground truth Python code containing a sequence of function primitives based on the language command. During inference, for each language command, we provide ChatGPT with the list of objects present in the SI Maps and generate Python code that refers to the specific instances involved in the language command.

In Figure \ref{fig:figure4}, we show a few examples of the Python executable code generated by ChatGPT for the given commands. ChatGPT successfully generates the correct executable code after prompting it with a few examples of language queries and corresponding ground truth Python executable code. To ground instances, our function primitives calls also include an instance parameter to handle instance-specific queries. The instance parameter is directly inferred from the language command by ChatGPT along with the object of interest. Overall, we define 23 function primitives for complex navigational maneuvers like moving between two objects, navigating to $n^{th}$ closest object, etc., and the essential turning and moving primitives.

\section{Experiments}

\subsection{Experimental Setup}
\label{exp_setup}

We showcase the effectiveness of our approach on multiple scenes from Matterport3D \cite{chang2017matterport3d} dataset in the Habitat \cite{savva2019habitat} simulator. Matterport3D is a commonly used dataset for evaluating the navigational capabilities of existing VLN agents in an indoor environment. The robot must maneuver in a continuous environment, performing navigational maneuvers specified by the natural language command. For top-view map creation, we collect 5,267 RGB-D frames from 5 different scenes and store the camera pose for each frame.

\textbf{Baseline: } We evaluate against a logical baseline where the semantic top-view maps from the VLMaps-based approach are separated into separate instances. If the objects in the environment are well separated, the semantic segmentation output should already contain the information required to separate different instances of similar objects by simply applying connected components. As a result, our baseline involves applying connected components over the VLMaps output. However, in realistic scenarios, different instances of the same object can be close to each other; for example: in a restaurant, chairs belonging to the same table are close to each other. In such a scenario, just computing connected components will not work, as multiple instances will get clubbed into a single instance. 

\textbf{Evaluation Metrics: } Like prior approaches \cite{huang23vlmaps,schumann-riezler-2022-analyzing,jain2022ground} in VLN literature, we use the gold standard \textbf{\textit{Success Rate}} metric, also known as \textbf{\textit{Task Completion}} metric to measure the success ratio for the navigation task. We compute the \emph{Success Rate} metric through human and automatic evaluations. For automatic evaluation, we use the ground truth environment map and compute the \emph{Success Rate} using a pre-defined heuristic where the navigation sub-goal is considered successful if we stop within a threshold distance of the ground truth object. For human evaluation, we verify if the agent ends up in a position desired according to the query.

\subsection{Evaluation Results}

In this section, we perform quantitative and qualitative comparisons of SI Maps against VLMaps and VLMaps with connected components. We compare the performance of each scene representation for the downstream language-based navigation task using the \emph{Success Rate} in table \ref{tab:success_rate}. We use the same function primitives for all the methods.

Human evaluation was done because of the observation made during a few queries where the agent ended up close to the target object, but it did not complete the task in the desired way. 

\begin{table}[h]
\centering
\renewcommand{\arraystretch}{1.5}
\begin{tabular}{lcc}
\hline
\multirow{2}{*}{Method} & \multicolumn{2}{c}{Success Rate}        \\ \cline{2-3} 
                        & Human Evaluation & Automatic Evaluation \\ \hline
VL Maps                 & 0.24               & 0.46               \\
VL Maps with CC         & 0.34               & 0.48               \\
SI Maps (K=5)           & \textbf{0.80}      & \textbf{0.88}      \\
SI Maps (K=9)           & 0.76               & \textbf{0.88}              \\ \hline

\end{tabular}

\caption{SI Maps outperform other baseline methods by significantly large margins on the \emph{Success Rate} metric. The best results are highlighted in \textbf{bold}.}
\label{tab:success_rate}
\end{table}

We observe that SI Maps exhibit a remarkable improvement in performance compared to other approaches. SI Maps achieve an impressive two-fold increase in success rate metric compared to 24\% obtained by VLMaps on human evaluation, demonstrating a substantial leap in the instance-specific goal navigation. Since VLMaps only contain semantic information, they fail on queries that refer to specific instances of an object, like ``navigate to the second counter". Our logical baseline, VLMaps with connected components, can handle some instance-specific queries, resulting in an incremental performance gain of 10\% for human evaluation than vanilla VLMaps. However, the success of this method is observed in scenes where neighboring instances of the same object have ample room between them. In contrast, real-life environments such as offices, restaurants, and hospitals often have objects in close proximity to each other. In these cases, instance-level information is essential for distinguishing between neighboring objects. SI Maps demonstrate robustness to object placement in the environment by directly utilizing the instance-level information provided by the instance segmentation model during the occupancy grid creation.


\subsection{Qualitative Results}

\begin{figure}[ht]
  \centering
  \includegraphics[width=0.85\columnwidth]{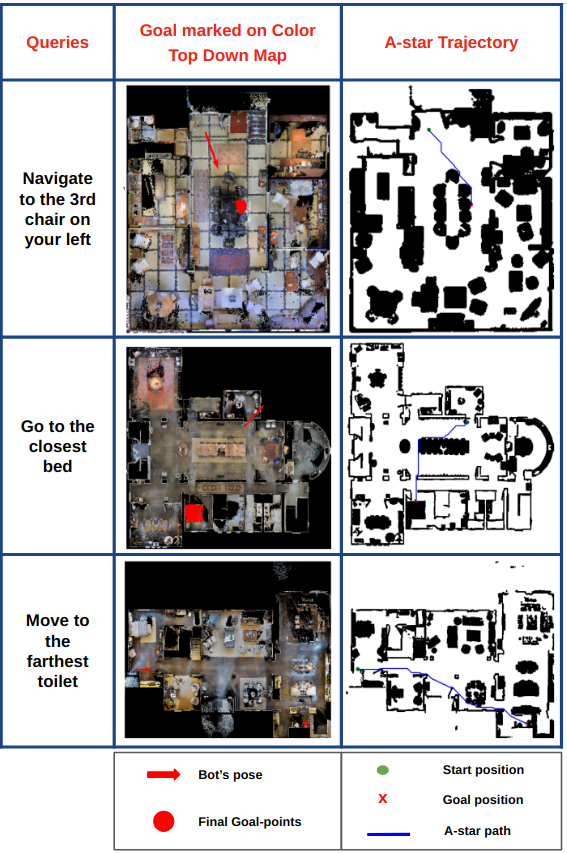}
  \caption{The above figure shows the agent in different scenes in a simulated environment with three different queries. Images on the top show the RGB top-down view map, along with the segmented goal object instance. The corresponding images on the bottom represent the path taken by the agent to reach the desired object from the initial location.}
  \label{fig:figure5}
\end{figure}

In this section, we showcase qualitative examples of our approach for the vision language navigation task. The results are illustrated in Figure \ref{fig:figure5} with the corresponding A-star trajectory using SI Maps for navigation. SI Maps allow navigating to specific instances in the scene based on their relative distance with respect to other objects (left, center) and direction-based specification in the global map (right). The downstream navigation, as a consequence of SI Maps, is agnostic to the starting pose and orientation of the agent in the environment. 

\begin{figure}[ht]
  \centering
  \includegraphics[width=\linewidth]{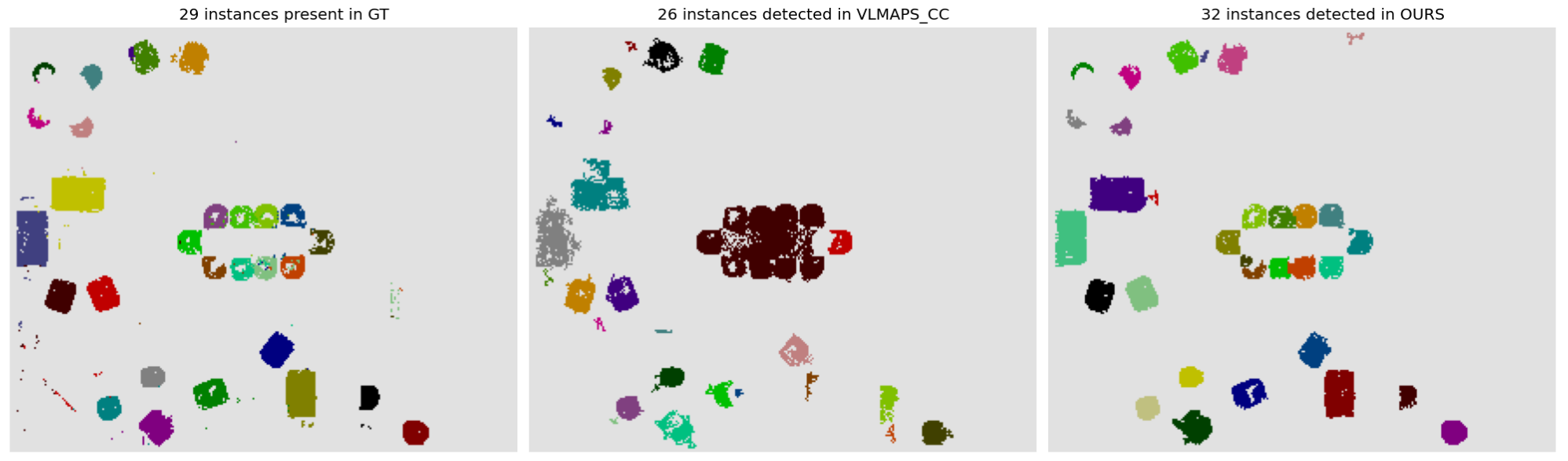}
  \caption{Qualitative example of the instance-level semantics captured by different methods for all the chairs in the environment. SI Maps clearly localize the different instances in the map.}
  \label{fig:figure6}
\end{figure}

We also show qualitative comparisons of different methods on the quality of instance-level top-view maps in Figure \ref{fig:figure6} for different seating objects (chair, couch, sofa) in the simulated environment. Our approach effectively captures the instance-level semantics of objects in the environment, recovering 32 instances out of 29 present in the map (with 3 extra noisy segments). In contrast, the baseline of VLMaps with connected components detects 26 instances, but most of them are noisy segments, and it merges several separate instances (for the same object in close proximity) into a single instance. Our results are particularly impressive in the middle region of the map, which corresponds to the dining area in the environment. Here, the chairs are in close proximity to each other, and the vanilla VLMaps approach fails when a particular instance of chair is queried. Similarly, applying connected components-based heuristics to separate instances is not enough, as the semantic segmentation masks of the chairs end up being connected with each other, resulting in multiple instances being merged.

The VLMaps-based approaches rely on alignment between per-pixel visual embeddings and linguistic feature embeddings, which can be sensitive to noise due to the unconstrained nature of the association. The benefit of our feature-embedding-free approach becomes evident as we directly constrain the occupancy grid creation with the instance segmentation masks. As a result, SI Maps have considerably less noise than derivative VLMaps approaches. Community detection further helps reduce noise by filtering out spurious communities formed due to noise, leading to a much cleaner map, which can also be observed in Figures \ref{fig:figure1}, \ref{fig:figure6}.

\section{Conclusion}

In this study, we introduce a novel instance-focused scene representation for indoor settings, enabling seamless language-based navigation across various environments. Our representation accommodates language commands that refer to specific instances within the environment. Furthermore, our map creation method is more memory-efficient, resulting in an impressive 128-fold decrease in storage, as it does not rely on high-dimensional feature embeddings for visual and linguistic modalities. Additionally, our approach demonstrates robustness in relation to object placement in the environment and is less vulnerable to noise than previous methods. We showcase the practicality of the proposed SI Maps using success rate and panoptic quality metrics. Future research could investigate 3D instance segmentation techniques to incorporate instance-level semantics into the occupancy grid creation process directly.

\section*{Acknowledgement}
We acknowledge iHub-Data IIIT Hyderabad for their support to this work.

\bibliography{root}
\bibliographystyle{ieeetr} 

\end{document}